%% file: root.tex
\documentclass[letterpaper, 10 pt, conference]{ieeeconf}  

\IEEEoverridecommandlockouts                              

\makeatletter
\let\NAT@parse\undefined
\makeatother

\input{packages}
\begin{document}

\author{Yiming Li$^{1,*}$, Zonglin Lyu$^{1,*}$, Mingxuan Lu$^2$, Chao Chen$^1$, Michael Milford$^3$, and Chen Feng\textsuperscript{1,\ding{41}}
\\
{\tt\small\url{https://ai4ce.github.io/CoVPR/}\vspace{-3mm}}
\thanks{* indicates equal contributions.}
\thanks{\ding{41} Corresponding author. This work is supported by NSF Grant 2238968.}
\thanks{$^{1}$Yiming Li, Zonglin Lyu, Chao Chen, and Chen Feng are with New York University,
Brooklyn, NY 11201, USA  {\tt\small\{yimingli, zl3958, cchen, cfeng\}@nyu.edu}}
\thanks{$^{2}$Mingxuan Lu is with Columbia University,
New York, NY 10027, USA  {\tt\small ml4799@columbia.edu}}
\thanks{$^{3}$Michael Milford is with the QUT Centre for Robotics, Queensland University of Technology, Brisbane, QLD 4000, Australia {\tt\small michael.milford@qut.edu.au}}
}

\title{\LARGE \bf Collaborative Visual Place Recognition}

\maketitle
\thispagestyle{empty}
\pagestyle{empty}
\input{parts/0-title-author-abstract}

\input{parts/1-intro}

\input{parts/2-related}
\input{parts/3-problem_model}

\input{parts/4-experiments}

\input{parts/5-Limitation}
\input{parts/6-conclusions}

{\small
\bibliographystyle{IEEEtran}  
\balance
\normalem
\bibliography{IEEEabrv,ref}
}
\end{document}

%% file: packages.tex
\usepackage{graphics} 
\usepackage{epsfig} 
\usepackage{empheq}
\usepackage{times} 
\usepackage{amsmath} 
\usepackage{amssymb}  
\usepackage{bm}
\usepackage{color}
\usepackage{bbding} 
\usepackage{diagbox}
\usepackage{comment}
\usepackage{float}
\restylefloat{table}
\usepackage{makecell}
\usepackage[square, comma, numbers,sort&compress]{natbib}
\usepackage{ulem} 
\usepackage{float}
\usepackage{booktabs}
\usepackage{multirow}
\usepackage{mathrsfs}
\usepackage[utf8]{inputenc}
\usepackage{graphicx}
\usepackage{pifont}
\usepackage{threeparttable}
\usepackage{caption}
\usepackage{setspace}
\newcounter{RNum}
\usepackage{balance}
\usepackage{multicol}
\usepackage{blindtext}
\usepackage[linesnumbered,ruled]{algorithm2e}
\usepackage{algorithmic}
\usepackage{amsfonts}
\usepackage{color}
\usepackage[table,dvipsnames]{xcolor}
\usepackage[pagebackref=true,breaklinks=true,colorlinks,bookmarks=false]{hyperref}

\hypersetup{
linkcolor=BrickRed
,citecolor=Green
,filecolor=Mulberry
,urlcolor=NavyBlue
,menucolor=BrickRed
,runcolor=Mulberry
,linkbordercolor=BrickRed
,citebordercolor=Green
,filebordercolor=Mulberry
,urlbordercolor=NavyBlue
,menubordercolor=BrickRed
,runbordercolor=Mulberry
}

\usepackage{etoolbox}  
\makeatletter
\patchcmd{\algorithmic}{\addtolength{\ALC@tlm}{\leftmargin} }{\addtolength{\ALC@tlm}{\leftmargin}}{}{}
\makeatother

%% file: parts/0-title-author-abstract.tex
\providecommand{\acronym}{CoVPR}
\providecommand{\realworld}{NYU-VPR-360}
\begin{abstract}
Visual place recognition (VPR) capabilities enable autonomous robots to navigate complex environments by discovering the environment's topology based on visual input. Most research efforts focus on enhancing the accuracy and robustness of single-robot VPR but often encounter issues such as occlusion due to individual viewpoints. Despite a number of research on multi-robot metric-based localization, there is a notable gap in research concerning more robust and efficient place-based localization with a multi-robot system. This work proposes collaborative VPR, where multiple robots share abstracted visual features to enhance place recognition capabilities. We also introduce a novel collaborative VPR framework based on similarity-regularized information fusion, reducing irrelevant noise while harnessing valuable data from collaborators. This framework seamlessly integrates with well-established single-robot VPR techniques and supports end-to-end training with a weakly-supervised contrastive loss. We conduct experiments in urban, rural, and indoor scenes, achieving a notable improvement over single-agent VPR in urban environments ($\sim$12\%), along with consistent enhancements in rural ($\sim$3\%) and indoor ($\sim$1\%) scenarios. Our work presents a promising solution to the pressing challenges of VPR, representing a substantial step towards safe and robust autonomous systems.
\end{abstract}


%% file: parts/1-intro.tex
\section{Introduction}

\begin{figure}[t]
\centering
    
    \includegraphics[width=\linewidth]{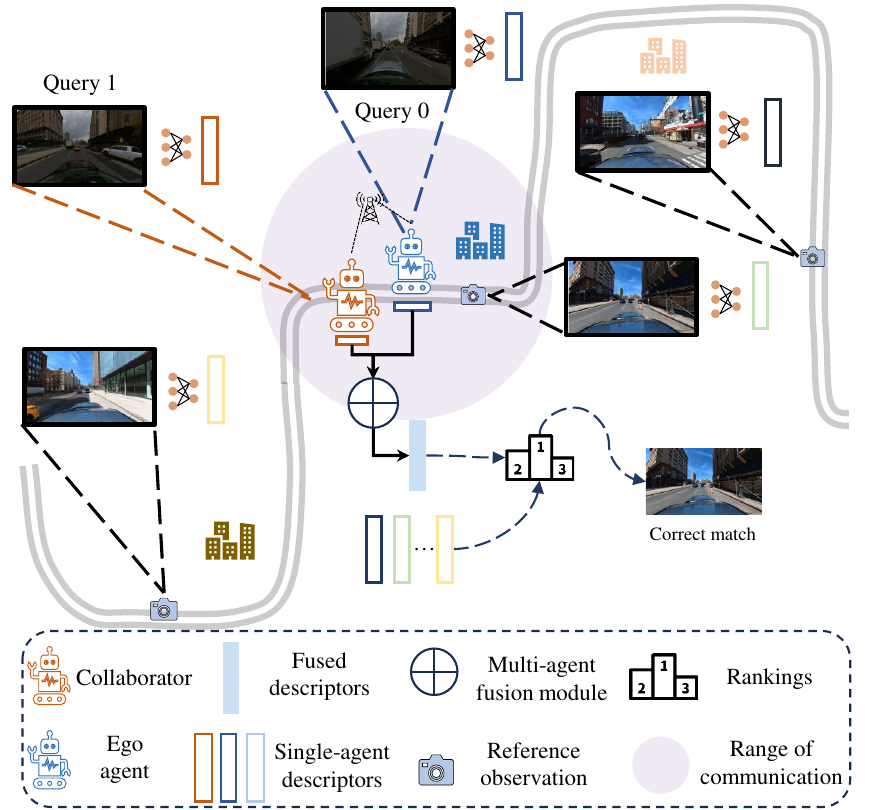}
    
    \caption{\textbf{Illustration of collaborative VPR.} The ego robot receives information from collaborators within a defined communication radius, generating a fused descriptor, which is then compared with single-robot descriptors in the database.}
    \label{general}
    \vspace{-5.5mm}

\end{figure}

Visual place recognition (VPR) is an important technology that enables autonomous robots to recognize and identify specific places or locations by analyzing visual input~\cite{schubert2023visual}. This capability proves crucial in situations where GPS signals may be unreliable or unavailable, as seen in urban canyons or indoor environments. To be more specific, VPR involves comparing real-time sensory streams, such as RGB images, to a reference database of previously seen images, allowing robots to understand the environment's topology and make informed decisions based on visual cues.

In practical applications, VPR systems face two major challenges: \textit{appearance variation} and \textit{limited viewpoint}. Appearance variation typically arises from changes over time and diverse observation angles, causing a query image to appear distinct from a reference image of the same location. Limited viewpoint occurs when there are insufficient recognizable landmarks, often due to occlusions, concealing essential details in the occluded portions of the query image. To address these challenges, recent research efforts~\cite{NetVLAD, patchnetvlad, transvpr, r2former, mixvpr} have employed deep architectures~\cite{cnn, ResNet, ViT, VGG} to generate robust latent features, which exhibit greater resilience against appearance variations compared to conventional approaches relying on handcrafted features~\cite{SIFT, SURF, ORB}. Some of these works~\cite{patchnetvlad, transvpr, r2former} have also developed matching strategies that exploit local keypoint features to enhance performance. To tackle the issues posed by limited viewpoints, prior studies have leveraged temporal information, as seen in sequential VPR~\cite{seqmatchnet, garg2021seqnet, Amusic}. However, it is worth noting that camera viewpoints in short video clips tend to be highly similar, providing insufficient complementary information.

A useful approach to address the issue of insufficient information for a single robot is multi-robot collaboration for better precision, robustness, and resilience. There is a body of research on \textit{collaborative metric localization}, which relies on information-sharing and coordination among robots or sensors to improve the accuracy and reliability of their position estimates~\cite{prorok2012low,wiktor2020collaborative}. However, there is limited research on \textit{collaborative topological localization}, where a team of robots collectively determines their positions within an environment through the \textit{collaborative recognition of distinctive landmarks or places}. Topological localization offers a notable advantage in certain scenarios, particularly when obtaining precise metric information is challenging or when the primary focus is on high-level navigation tasks~\cite{he2023metric}. It exhibits reduced sensitivity to sensor noise and environmental changes compared to purely metric-based approaches. 
 
In this paper, we initiate the first study of \textit{collaborative visual place recognition (\acronym)} in which multiple robots located in nearby areas share information to achieve collaborative topological localization, as shown in Fig.~\ref{general}. In addition to the novel problem formulation, we develop an effective and efficient weakly-supervised collaborative learning method for~\acronym~based on similarity-regularized information fusion, which can leverage the complementary information from other robots while minimizing the impact of irrelevant noise. Meanwhile, our method seamlessly integrates with established VPR techniques, such as NetVLAD~\cite{NetVLAD}, and can be trained end-to-end using a weakly-supervised contrastive objective. To validate its effectiveness, we utilize the Ford multi-AV dataset~\cite{Ford} and also collect a new collaborative driving dataset in New York City (NYC MV dataset). Beyond outdoor scenarios, we also evaluate our approach in indoor environments with a simulated Gibson multi-robot dataset~\cite{gibson}. Our contributions are threefold:

\begin{itemize}
    \item We formulate \textit{collaborative visual place recognition} (\acronym), a meaningful and challenging task for the vision and robotics community.
    \item We develop a novel and effective CoVPR method based on similarity-regularized fusion which is widely applicable to learning-based single-robot VPR methods.
    \item We provide a comprehensive CoVPR benchmark that includes publicly available datasets and a dataset we collected ourselves, encompassing a wide range of scenarios, from outdoor to indoor environments.
\end{itemize}

%% file: parts/2-related.tex
\section{Related Works}

\label{sec:related_works}

\textbf{Visual place recognition.} 
Traditional visual place recognition methods aggregate handcrafted features, such as SIFT~\cite{SIFT} and SURF~\cite{SURF}, into global descriptors~\cite{BOW,VLAD,vlad2}. While these handcrafted features offer scale invariance, their expressiveness pales in comparison to features extracted by deep learning models~\cite{ResNet,ViT}. To bridge the gap between VPR and deep learning, NetVLAD~\cite{NetVLAD}, combined with triplet ranking loss, has paved the way for end-to-end VPR models and is followed by subsequent efforts for further advancements~\cite{CRN,adageo,Attention-Aware,spatial,VLAAD,APPSVR,chen2022self}. Other approaches like R-MAC~\cite{rmac}, GeM~\cite{GeM}, GAP~\cite{GAP}, and GMP~\cite{GMP} have introduced non-trainable pooling layers compatible with deep learning, resulting in lightweight models. Beyond pooling layers, MixVPR enhances feature richness through channel-wise and feature-wise mixtures using MLPs~\cite{mixvpr}. Recent advancements in VPR include re-ranking modules that leverage RANSAC on geometric information from local patch descriptors, further boosting performance~\cite{patchnetvlad,transvpr}. $R^2$Former streamlines training and re-ranking modules into a single framework~\cite{r2former}, eliminating the intensive computation of RANSAC. In addition to image-level VPR, several research explores video-level VPR with novel training and matching designs~\cite{garg2021seqnet,seqmatchnet,Amusic}. Despite the wealth of research in various VPR aspects, none of these prior efforts have ventured into the domain of multi-robot collaborative VPR.

\textbf{Multi-robot mapping and localization.} Simultaneous localization and mapping (SLAM) is a vital research area in robotics, where robots concurrently construct maps of their environments while determining their own locations~\cite{SLAM,ding2019deepmapping,chen2023deepmapping2}. Extensive research efforts have been dedicated to multi-agent SLAM~\cite{MULTIROBOTSLAM,MULTIPOSEGRAPH,ccmslam,moarslam,MAANS}. \cite{MULTIROBOTSLAM} and~\cite{MULTIPOSEGRAPH} leverage relative range sensor measurements to estimate the global state.~\cite{ccmslam,moarslam} adopt centralized architectures for map fusion.~\cite{MAANS} incorporates Transformers~\cite{transformer} to integrate spatial and inter-team information. Additionally, \cite{prorok2012low} and \cite{Spasojevic2023ActiveCL} achieve collaborative metric localization using filtering-based methods. These efforts predominantly concentrate on multi-robot collaboration within the context of metric mapping and localization, with the goal of constructing a global map by fusing multiple local maps or improving localization through additional measurements from various robots. Nevertheless, online collaborative place-based topological localization remains unexplored.

\textbf{Collaborative perception.} Multi-robot collaborative perception, where multiple robots collaborate by sharing their sensor observations to enhance perception capabilities, has emerged as a promising solution to tackle the issue of limited viewpoints in single-robot perception, especially in large-scale driving scenes~\cite{li2022v2x,xu2022opv2v}. Existing research primarily focuses on collaborative object recognition, such as 3D object detection based on either 3D LiDAR point clouds~\cite{li2021learning} or 2D RGB images~\cite{hu2023collaboration}. Additionally, several studies delve into collaborative semantic segmentation for multiple drones~\cite{liu2020who2com, zhou2022multi} or vehicles~\cite{xu2022cobevt, li2022multi}. Researchers have addressed various aspects of collaborative perception, including communication efficiency~\cite{hu2022where2comm,li2021learning}, uncertainty estimation~\cite{su2022uncertainty,su2023collaborative}, adversarial robustness~\cite{tu2021adversarial,amongus}, and domain adaptation~\cite{xu2023bridging}. However, there is a noticeable absence of research on collaborative place recognition, which could fundamentally resolve the issue of limited viewpoints in single-robot VPR. In collaborative object recognition, multiple robots share the same ground truth 3D object locations. However, collaborative place recognition presents a unique challenge: each robot operates with distinct ground truth locations, which could result in noisy information sharing, as shared messages may indicate different places or locations.

%% file: parts/3-problem_model.tex
\section{Methodology}

\label{sec:methodology}

\begin{figure*}[t]

\centerline{\includegraphics[width=0.95\linewidth]{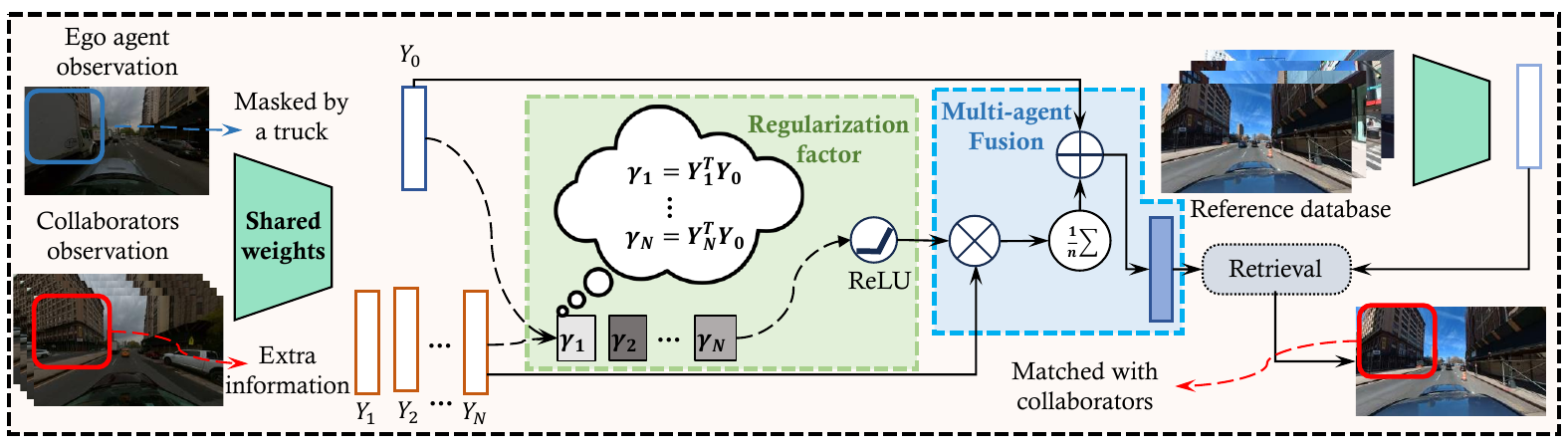}}
\caption{\textbf{Overall architecture of our methods.} Each image is processed by a shared feature extractor (in green) to obtain a descriptor. Subsequently, the regularization factor is computed, and descriptors from different agents are fused into a collaborative descriptor. The darkness of $\gamma$ represents its values, with darker shades indicating larger values.}

\label{archi}
\vspace{-3mm}
\end{figure*}

In this section, we begin by outlining the problem of CoVPR and subsequently introduce our method. The core concept centers on an ego agent's quest to determine its geographical location based on visual cues, aided by nearby agents (collaborators). In this configuration, although we lack precise knowledge of the collaborators' exact locations, we do have access to their visual observations when they are located within a certain communication range. The ego agent effectively aggregates information from both other agents and its own data to conduct place recognition. The overall architecture of our model is depicted in Fig.~\ref{archi}.

\subsection{Problem Definition} 
\label{def}
\textbf{Notations.} We index the ego agent as $0$ and collaborators as $1,2,...,N$. Moreover, we denote the ego agent query image as $I_0$, collaborators query images as $I_1,...,I_N$, and a reference image as $I_D$. The single-agent VPR model, which takes images as input and descriptors as output, is denoted as $F:\mathcal{I}\mapsto \mathbb{R}^d$ where $\mathcal{I}$ is the image space. Descriptors are denoted as $Y_D$ and $Y_i$ for $i = 1,...,N$. The multi-agent fusion module, which aggregates descriptors extracted from different agents into a single descriptor, is denoted as $G$.

\textbf{Constraints.} The multi-agent fusion module $G$ is subject to several constraints to ensure its proper functioning.

\begin{enumerate}
    \item \textbf{Consistency.} If a set of multi-agent query images $\{I_0, \ldots, I_N\}$ are identical, the resulting multi-agent descriptor should be identical to the individual single-agent descriptors: $G(Y_0, \ldots, Y_N) = Y_i$ for $i = 0, \ldots, N$. In other words, the descriptor of an image should remain consistent, whether computed in the context of single-agent or multi-agent.

    \item \textbf{Permutation.} The output descriptors should remain invariant when the order of collaborators is permuted. However, swapping the ego agent with a collaborator will result in a different descriptor.
\end{enumerate}

\textbf{Challenges.} As an unexplored research area, CoVPR presents several noteworthy challenges. 
\begin{enumerate}
    \item \textbf{Dispersed collaborators.} Collaborators can be widely dispersed, which significantly differs from the relatively small displacement within video clips in sequential VPR setups. This dispersion poses a unique challenge, as techniques designed for sequential setups may not be directly applicable to CoVPR. Sequential techniques often assume a single location per video clip, which violates the \textit{permutation constraint}. Moreover, the relatively large distances among agents in CoVPR potentially introduce noise to the ego agent's data, which can adversely affect performance. Coping with this increased distance and mitigating its impact is crucial for accurate place recognition.
    \item \textbf{Asymmetry in query and database images.} A set of CoVPR queries consists of multiple images, while a database reference typically contains one single image. This asymmetry poses an unique challenge to the \textit{consistency constraint}. Trainable fusion modules often apply randomly initialized weights $W$ to a descriptor $Y$. Ensuring $WY = Y$ to meet the \textit{consistency constraint} is challenging due to the random nature of weight initialization. Therefore, designing fusion modules that account for this asymmetry is essential to effectively address this challenge.
\end{enumerate}

\subsection{VLAD}
Vector of Locally Aggregated Descriptors (VLAD)~\cite{VLAD,vlad2} stands as a prominent pooling method for retrieval. It operates by computing residuals between descriptors and cluster centroids. Typically, descriptors are pre-computed, and cluster centroids are obtained by unsupervised learning. The residuals are then aggregated and normalized to generate a feature vector (global descriptor). Given M descriptors of an image $\mathbf{x}_1,...,\mathbf{x}_M \in \mathbb{R}^d$ and K cluster centroids $\mathbf{c}_1,...,\mathbf{c}_K \in \mathbb{R}^d$, VLAD yields K distinct vectors $\mathbf{V}_1,...,\mathbf{V}_K \in \mathbb{R}^d$. Each $\mathbf{V}_k$ is calculated as follows:

\begin{equation}
    \label{eq:VLAD}
    \mathbf{V}_k = \sum_{i = 1}^{M}\mathbf{\alpha}_k(\mathbf{x}_i)(\mathbf{x}_i - \mathbf{c}_k) \ ,
\end{equation}

\begin{equation}
\label{VLADcondition}
  \alpha_k(\mathbf{x}_i) =
    \begin{cases}
      1 & \text{if $||\mathbf{x}_i - \mathbf{c}_k||<||\mathbf{x}_i - \mathbf{c}_j|| \text{ }\forall j \neq k$}\\
      0 & \text{otherwise}  \ .
    \end{cases}       
\end{equation}

Subsequently, $\mathbf{V}_1,...,\mathbf{V}_K$ undergo an independent $L^2$ normalization (intra-normalization), followed by concatenation and a global $L^2$ normalization to yield a VLAD vector. Image retrieval is based on the $L^2$ distance between VLAD vectors.

\subsection{NetVLAD}
The computation of $\mathbb{\alpha}_k(\mathbf{x}_i)$ in VLAD is non-differentiable and thus incompatible with Gradient Descent. To leverage Deep Learning, NetVLAD~\cite{NetVLAD} proposes a differentiable computation approximating the functionality of VLAD, where $\mathbb{\alpha}_k(\mathbf{x}_i)$ is replaced by $\bar{\mathbb{\alpha}}_k(\mathbf{x}_i)$ such that:

\begin{equation}
    \label{eq:NetVLAD}
    \bar{\mathbf{\alpha}}_k(\mathbf{x}_i) = \frac{e^{-\alpha||\mathbf{x}_i - \mathbf{c}_k||^2}}{\sum_{k'}e^{-\alpha||\mathbf{x}_i - \mathbf{c}_{k'}||^2}} \ .
\end{equation}

Expanding $L^2$ norm:

\begin{equation}
    \label{eq:NetVLAD2}
    \bar{\mathbf{\alpha}}_k(\mathbf{x}_i) = \frac{e^{w_k^T\mathbf{x}_i + b_k}}{\sum_{k'}e^{w_{k'}^T\mathbf{x}_i + b_{k'}}},
\end{equation}
\begin{equation}
    \label{eq:NetVLAD3}
    w_k = 2\alpha\mathbf{c}_k,b_k = -\alpha||\mathbf{c}_k||^2.
\end{equation}

Replacing pre-computed features with features extracted by convolutional neural networks (CNNs), NetVLAD is end-to-end trainable with the above adjustment.

\subsection{Similarity-Regularized Fusion}

To address the challenges mentioned earlier, we introduce a method known as \textit{similarity-regularized fusion} for aggregating information from all agents. For a feature extractor that generates global descriptors, denoted as $Y_0, \ldots, Y_N$ for $N$ collaborators, the collaborative descriptor is defined as:

\begin{equation}
    \label{eq:collaborative}
    Y = \sigma\left(Y_0 + \frac{1}{N}\sum_{n = 1}^{N}\gamma_nY_n\right),
\end{equation}

\begin{equation}
    \label{gamma}
    \gamma_n = \max(Y_0^TY_n,0),\text{ }\sigma(x) = \frac{x}{||x||^2}.
\end{equation}

A naive approach to aggregating information from the ego agent and collaborators is average pooling. However, it is important to note that this method violates the second part of the \textit{permutation constraints} in Section \ref{def}. Moreover, due to the distances between the ego agent and collaborators, this naive operation might degrade performance. To address this, we introduce a $\gamma_n$ term, which regulates the influence of $Y_n$ on $Y_0$ while adhering to the \textit{permutation constraints}. This operation introduces a minor perturbation to $Y_0$, slightly orienting it towards $Y_n$, with the degree of rotation controlled by $\gamma_n$. Additionally, we apply the scaling factor $\frac{1}{N}$ to emphasize the contribution of the ego agent, ensuring that it is not overshadowed by a large number of collaborators.

\textbf{Constraint satisfaction.}
\begin{enumerate}
    \item \textbf{Consistency.} Our method upholds the consistency constraint because $G(Y_0, Y_1, \ldots, Y_N) = \sigma((N+1)Y_i) = Y_i$. This ensures that an image's descriptor remains consistent, whether processed by the multi-agent fusion module or considered as a single-agent descriptor.

    \item \textbf{Permutation.} The permutation constraint relies on the permutation-equivariant properties of dot products. The subsequent summation operation ensures invariance. Furthermore, the coefficient $\gamma$ is determined by the value of $Y_0$, so swapping the ego agent and collaborator results in distinct descriptors.
\end{enumerate}

\textbf{Cluster-wise variant.} In the context of NetVLAD pooling, our method can be applied independently to each intra-normalized cluster descriptor. Consequently, we calculate $\gamma$ separately. The cluster-wise variant is expressed as:

\begin{equation}
    \label{eq: cluster collaborative}
    V_k = \sigma\left(V_{0,k} + \frac{1}{N}\sum_{n = 1}^{N}\gamma_nV_{n,k}\right),
\end{equation}

\begin{equation}
    \label{gamma2}
    \gamma_n = \max(V_{0,k}^TV_{n,k},0),\text{ }Y = \sigma([V_1, \ldots, V_K]).
\end{equation}

\textbf{Implicit trainable parameters.} While $Y_0^TY_i$ does not seem to contain trainable parameters, it influences the gradient direction, rather than being a constant scalar.

\textbf{Initialization from single-agent.} Our method relies on effective similarity scores. Therefore, initializing a multi-agent model with the weights of a trained single-agent model benefits stability and convergence.

%% file: parts/4-experiments.tex
\section{Experiments}

\begin{figure}[t]
\centering
    
    \includegraphics[width=1\linewidth]{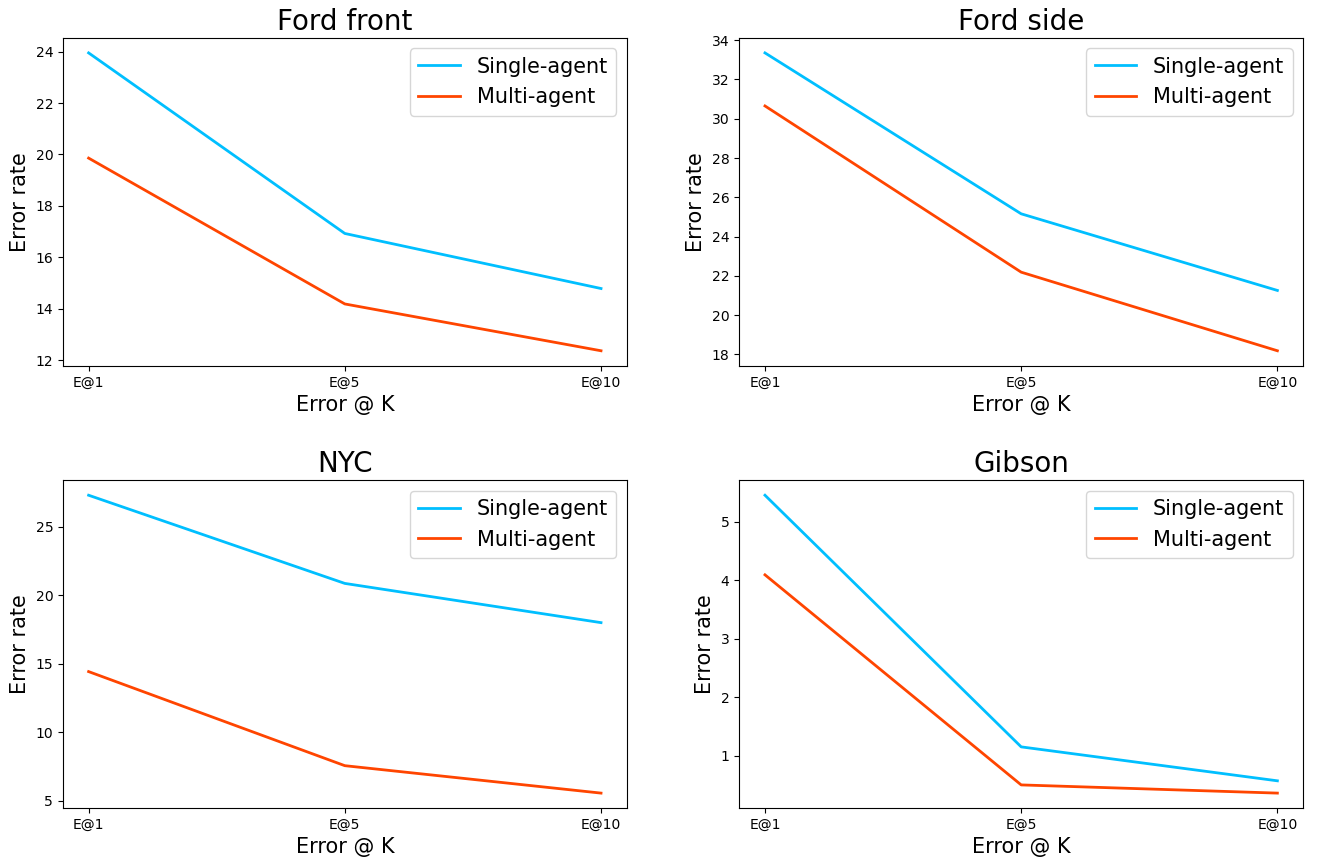}
    
    \caption{\textbf{Error rate visualization.} For simplicity, we denote the three datasets as Ford, NYC, and Gibson.}
    \label{CoVPR}
\end{figure}

\begin{figure*}[t]

    \centerline{\includegraphics[width=0.85\linewidth]{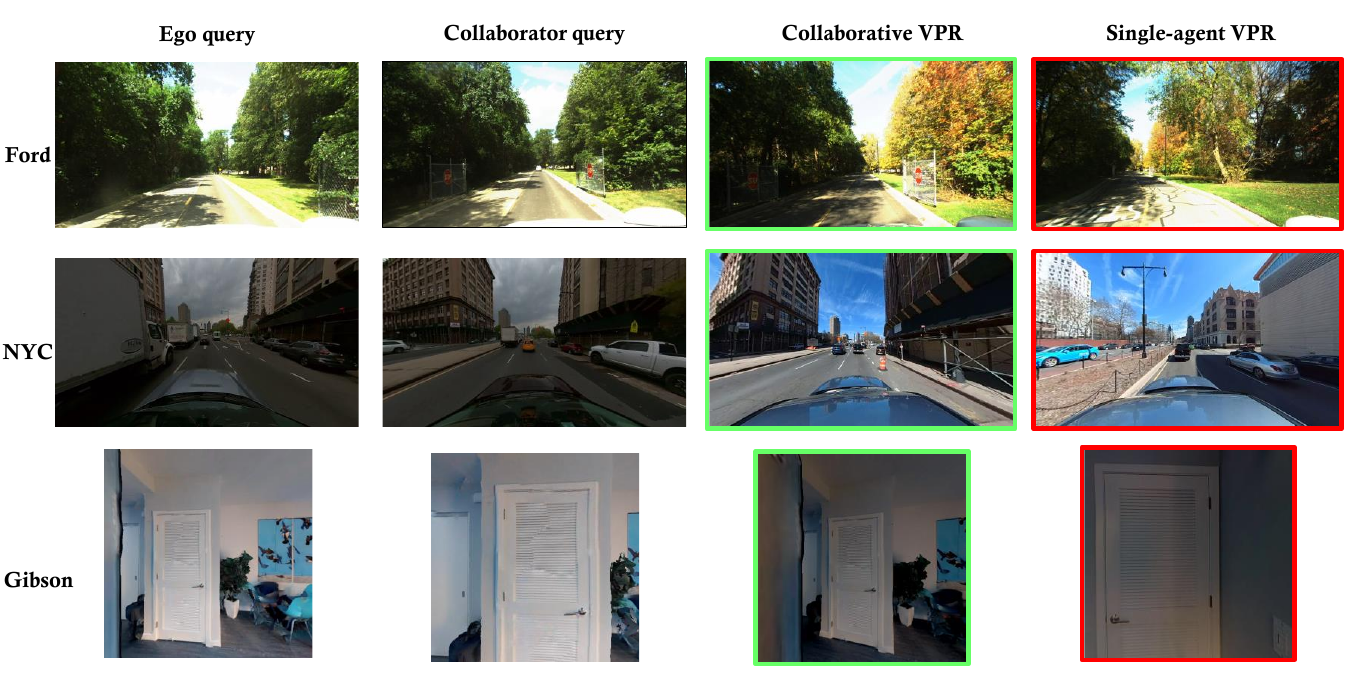}}

    \caption{\textbf{Qualitative examples of our method.} Incorrect retrievals are highlighted with \textcolor{red}{red rectangles}, while correct ones are indicated with \textcolor{Green}{green}. The observed improvements are often attributed to the additional information provided by collaborators.}

    \label{example}
\vspace{-3mm}
\end{figure*}

We evaluate our method using NetVLAD~\cite{NetVLAD} on three datasets: the Ford multi-AV Dataset~\cite{Ford}, the NYC MV dataset, and the Gibson multi-robot dataset~\cite{gibson}.

\textbf{Ford multi-AV dataset.} This dataset comprises outdoor scenes in rural Michigan. Evaluation includes both front and right-side views. We use 480 queries from residential scenes for training and 430 queries from university and vegetation scenes for testing. The maximum distance between the ego agent and collaborators is 5 meters.

\textbf{NYC MV dataset.} Captured in New York City, this dataset consists of images from the dense urban environment. The train-test split is based on GPS information to ensure disjoint locations. We use 207 queries for training and 140 for testing. The maximum distance between the ego agent and collaborators is 10 meters.

\textbf{Gibson multi-robot dataset.} The Gibson multi-robot dataset contains indoor scenes simulated with Habitat-sim~\cite{habitat19iccv}. We selected 350 queries for training and 279 queries for testing with front views. The training set and the test set do not share the same room, and the maximum distance between the ego agent and collaborators is 1 meter.

\subsection{Experimental Setup}

\textbf{Training details.} We employ ResNet18~\cite{ResNet} pretrained on ImageNet1K~\cite{imagenet} as the backbone for NetVLAD. Images are resized to $400\times200$ for outdoor scenes and $200\times200$ for indoor scenes. The number of clusters is set to 32, and we train the models up to 30 epochs with the Adam optimizer~\cite{Adam} five times. The learning rate is set to be 1e-4 for outdoor scenes and 1e-5 for indoor scenes, halved every five epochs. Other details follow the NetVLAD paper~\cite{NetVLAD}. As NetVLAD has a fixed initialization, we maintain consistent initialization in the multi-agent model, ensuring that all models are subjected only to sampling randomness during negative sample selection.

\begin{table}[t]
    \resizebox{\columnwidth}{!}{%
    \caption{\textbf{Ablation study on re-ordering.} We observe a performance drop when applying the re-ordering method. The best performances are \textbf{in bold}. \vspace{-3mm}}
    \label{Reorder}
    
    \begin{tabular}{c|cc|cc|cc}
        \Xhline{4\arrayrulewidth}
        \multicolumn{7}{c}{Dataset} \\
        \hline
         \multirow{2}{*}{Recall} & \multicolumn{2}{|c|}{Ford Front} & \multicolumn{2}{|c|}{Ford Side} & \multicolumn{2}{|c}{NYC}\\
        \cline{2-7}
        
         &{\footnotesize Single-agent} & {\footnotesize Re-ordering} & {\footnotesize Single-agent} & {\footnotesize Re-ordering} & {\footnotesize Single-agent} & {\footnotesize Re-ordering}\\
        \hline
        R@1 & \textbf{76.05} & 76.04  & \textbf{66.65} & 65.53 & \textbf{72.71} & 67.43 \\
        R@5 & 83.07 & \textbf{83.91}  & \textbf{74.84} & 74.51 & 79.14 & \textbf{80.14} \\
        R@10 & 85.21 & \textbf{86.42}  & \textbf{78.74} & 78.04 & 82.00 & \textbf{83.57} \\
        \Xhline{4\arrayrulewidth}
    \end{tabular}
    }
    
\end{table}

\textbf{Evaluation metric.} 
We use the recall rate (recall at K) as our primary evaluation metric, which is defined as the ratio of correct retrievals to the number of queries. In the context of recall at K, a correct retrieval occurs when the geographical distance between the retrieved image and the query is less than $M$ meters. For a better illustration, we report the error rate at K, calculated as 1 - recall. We set $M$ to be 20 meters in outdoor scenes and 1.5 meters in indoor scenes. Notably, queries from multiple agents have distinct coordinates, so we use the coordinate of the ego agent for evaluation. During the evaluation, the single-agent NetVLAD serves as the baseline model, and the recall rate is averaged across five trained models to reduce sampling variance. 

\subsection{Quantitative Results} 

\begin{figure}[t]
    \centering
    \resizebox{1\columnwidth}{!}{%
    \includegraphics[width=1\linewidth]{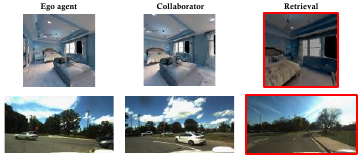}
    }
    \caption{\textbf{Qualitative examples of failure cases.} Failure cases are usually caused by large pose differences, potentially adding noises to the ego agent.}

    \label{example2}
    
\end{figure}

Our method is able to achieve more than $50\%$ reduction in error rate in the NYC MV dataset and the Gibson multi-robot dataset, as well as more than $10\%$ reduction in the Ford multi-AV dataset. Further details are presented in Fig.~\ref{CoVPR}. Moreover, we conduct an analysis of cases where collaboration either enhances or hinders the baseline performance. In general, collaboration proves more effective in some challenging scenarios. When the ego agent's camera view is obstructed, when the ego agent is navigating through dense vegetation that is difficult to recognize, and when the ego agent is looking at a closet without recognizable items, collaboration significantly enhances performance. However, if a large rotation exists between the view of the ego agent and collaborators, it is possible for collaboration to harm performance. To provide visual context, sample images illustrating performance improvement are presented in Fig.~\ref{example}, while some failure cases are shown in Fig.~\ref{example2}.

\textbf{Communication costs.} In the defined setup, the VLAD vector has a size of 16,384, approximately a $72\times 72$ RGB image. We can further compress this feature representation using an autoencoder for real-world deployment.

\begin{table}[t]
    \resizebox{\columnwidth}{!}{%

    \caption{\textbf{Ablation study on average pooling and cross-evaluation.} In the table, \textit{single-single} represents training and testing in single-agent cases, and \textit{single-multi} and \textit{multi-single} are similarly defined. The best performances are \textbf{in bold} for each configuration.\vspace{-3mm}}

    \label{Pool}
    
    \begin{tabular}{c|c|c|c|c}
        \Xhline{4\arrayrulewidth}
        \multicolumn{5}{c}{Dataset} \\
        \hline
         \multirow{2}{*}{Recall} & \multicolumn{4}{|c}{Ford Front}\\
        \cline{2-5}
        
         &Single-single & Average Pooling & Multi-single & Single-multi \\
        \hline
        R@1 & 76.05 & 77.72  & \textbf{78.93} & 76.37 \\
        R@5 & 83.07 & 84.10  & \textbf{84.70} & 83.81  \\
        R@10 & 85.21 & 86.42  & \textbf{86.60} & 86.37 \\
        \Xhline{4\arrayrulewidth}
    \end{tabular}
    }
\end{table}

\subsection{Ablation Study}

\textbf{Re-ordering.} We implement a straightforward re-ordering technique, inspired by prior work in the Sequential setup \cite{Amusic,seqmatchnet,garg2021seqnet}. In this technique, we treat the ego agent and collaborators as if they share the same ground-truth location, so their ranking scores are summed to a cumulative score. However, this approach may not yield optimal results because the agents are distributed across different locations. Quantitative results are presented in Table \ref{Reorder}. Notably, there is an obvious decrease in performance, particularly in the side-view scenario, which is highly sensitive to translations.

\textbf{Average pooling and cross-evaluation.} We employ average pooling to directly combine all VLAD vectors, taking the average and subsequently re-normalizing the result. Additionally, we perform cross-evaluation, wherein we train a collaborative model but evaluate it in a single-agent setting (multi-single), and vice versa (single-multi). Although average pooling does improve performance to some extent, it does not achieve the same level of improvement as our proposed method. In the context of cross-evaluation, we observe that the performance surpasses the baseline, and multi-single achieves the best result. This observation suggests that our fusion module aids a VPR model in converging towards a more optimal solution (multi-single), and the integration of different images provides additional information (multi-multi). Quantitative details are provided in Table \ref{Pool}.

\begin{table}[t]
    \resizebox{\columnwidth}{!}{%
    \caption{\textbf{Ablation study on the impact of distance.} When the distance is increased to 8m, other methods start harming the performance, but our method still boosts the performance by a significant margin. The best performances at each distance are \textbf{in bold}.\vspace{-3mm}}
    \label{distance_other}
    
    \begin{tabular}{c|c|cc|cc|cc}
        \Xhline{4\arrayrulewidth}
        \multicolumn{8}{c}{Method} \\
        \hline
         \multirow{2}{*}{Recall}  & \multirow{2}{*}{Single-agent}& \multicolumn{2}{|c|}{ Re-ordering} & \multicolumn{2}{|c|}{Average Pooling}& \multicolumn{2}{|c}{Ours} \\
        \cline{3-8}
        
        & &5m & 8m & 5m & 8m & 5m & 8m\\
        \hline
        R@1 & 76.05 & 76.04 & 75.58  & 77.72 & 75.30 & \textbf{80.14} & \textbf{79.77} \\
        R@5 & 83.07 & 83.91 & 83.58  & 84.10 & 80.93 & \textbf{85.81} & \textbf{84.28}\\
        R@10 & 85.21 & 86.42 & 86.28  & 86.42 & 84.39 & \textbf{87.63} & \textbf{86.42}\\
        \Xhline{4\arrayrulewidth}
    \end{tabular}
    }
    
\end{table}

\begin{figure}[t]
    \centering
    \resizebox{0.9\columnwidth}{!}{%
    \includegraphics[width=1\linewidth]{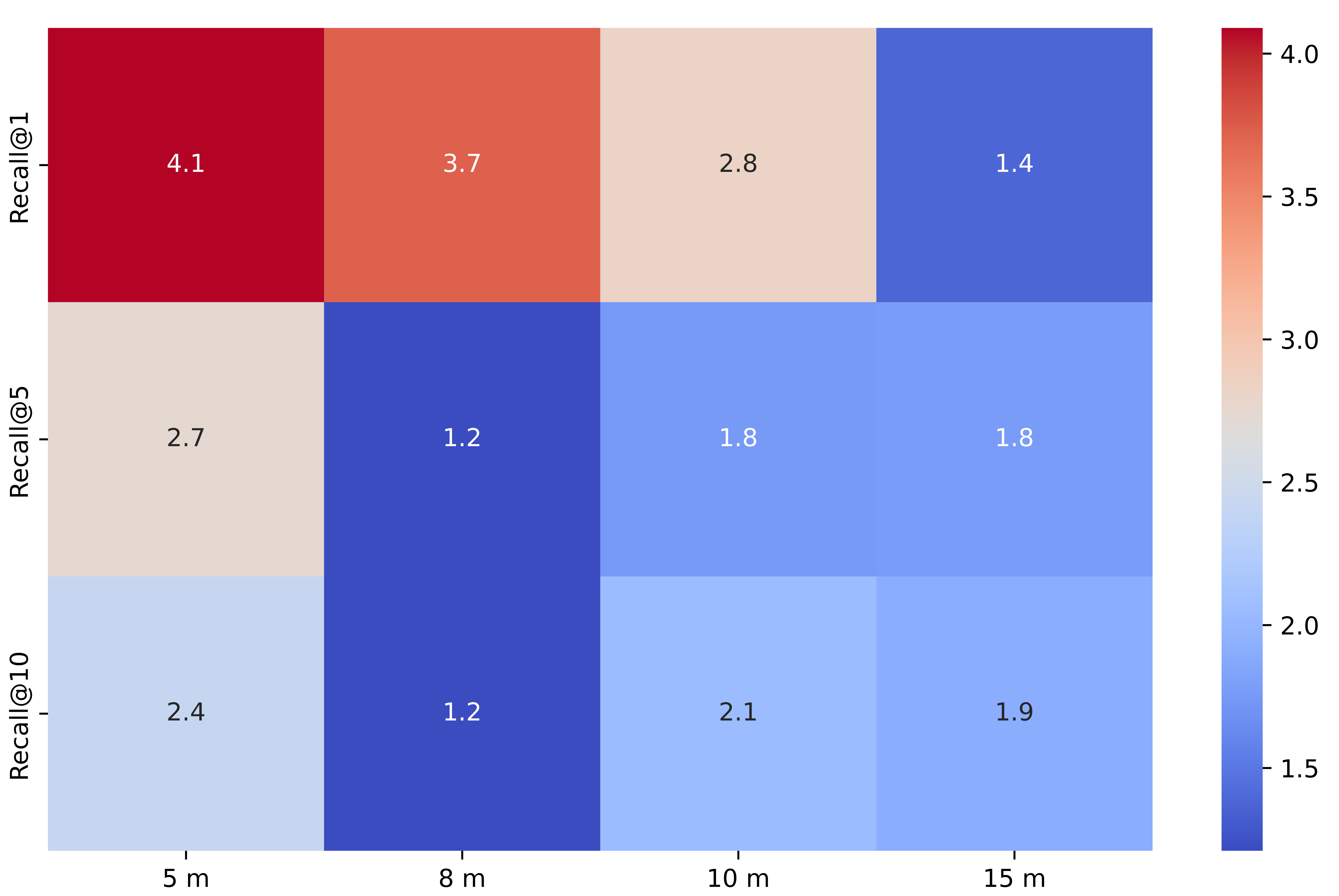}
    }
    \caption{\textbf{Performance gain under different distances between the ego agent and collaborators.} Our method remains relatively stable when distance is changing, except that performance gain on recall@1 is slightly decreasing.}
    \label{heatmap}
    \vspace{-3mm}
    
\end{figure}

\textbf{Impact of Distance.} An intriguing aspect of our investigation concerns the influence of the distance between the ego agent and collaborators on performance. Notably, we examine how our proposed method, average pooling, and re-ordering are affected as we extend distances from 5 meters to 8 meters. The results, as depicted in Table \ref{distance_other}, reveal that these techniques tend to degrade VPR performance under such conditions, rendering them less practical for real-world applications. However, our proposed method exhibits robustness even when distances are increased to 15 meters, suggesting its superior adaptability to real-world scenarios. While the performance gain in terms of recall at 1 diminishes with greater distances, it remains positive, and recall at 5 and 10 remains almost unchanged. This comprehensive evaluation of our method is illustrated in Fig.~\ref{heatmap}.

%% file: parts/5-Limitation.tex
\section{Limitations and Future Works}

While presenting a novel multi-agent collaboration mechanism in CoVPR, we acknowledge certain limitations and provide directions for future research.

\textbf{Limitations.} Although we have developed a method for more than two agents collaborating, we only conducted experiments with two agents. Practical constraints, such as the high cost of data collection, limited our experimentation.

\textbf{Future Works.} It would be interesting to develop a multi-agent collaboration module highly compatible with patch descriptors or to design an explicitly trainable collaboration module. Recent works have introduced re-ranking modules for VPR, but most of them involve searching for mutual nearest neighbors of patch descriptors, potentially overlooking additional information from collaborators. Furthermore, our method does not include explicit trainable weights in the fusion module. It would be intriguing to develop a trainable module while satisfying the \textit{consistency constraint}.

%% file: parts/6-conclusions.tex
\section{Conclusion}\label{sec:conclusion}
In this study, we introduce and formulate a novel CoVPR problem to tackle the issue of limited viewpoint and provide an innovative approach for multi-agent collaborative VPR. Our pioneering multi-agent collaboration mechanism largely enhances VPR performance in complex urban environments and still consistently enhances performance in rural and indoor environments. This versatile framework can be easily applied in single-robot VPR models such as NetVLAD, ensuring consistent and robust results. Additionally, it incurs a small communication overhead and operates efficiently in real-world applications. We believe our work can promote the development of multi-robot topological localization.